\documentclass{llncs}
\usepackage{enumerate}   

\usepackage{amsmath}
\usepackage{graphicx}
\usepackage{makeidx}  % allows for indexgeneration
\usepackage{etoolbox} % for "\patchcmd" macro
\usepackage{kotex} % for use special character \&
\usepackage{cite}
\usepackage{amssymb}  % assumes amsmath package installed
\usepackage{bm}
\makeatletter
\patchcmd{\ps@headings}{\rlap{\thepage}}{}{}{}
\patchcmd{\ps@headings}{\llap{\thepage}}{}{}{}
\makeatother
\begin{document}
%
%\frontmatter          % for the preliminaries
%
%\pagestyle{headings}  % switches on printing of running heads
%\addtocmark{Hamiltonian Mechanics} % additional mark in the TOC

%\tableofcontents
%
\mainmatter              % start of the contributions
\title{Sparse Identification of Nonlinear Dynamics-based Model Predictive Control \\ for Multirotor Collision Avoidance}
\titlerunning{Hamiltonian Mechanics}  % abbreviated title (for running head)
\author{Jayden Dongwoo Lee\inst{1} \and Youngjae Kim\inst{1}
Yoonseong Kim\inst{2} \and \\ Hyochoong Bang\inst{1}}
\authorrunning{Lee et al.} % abbreviated author list (for running head)
%
% %%%% list of authors for the TOC (use if author list has to be modified)
% \tocauthor{Ivar Ekeland, Roger Temam, Jeffrey Dean, David Grove,
% Craig Chambers, Yuseop Shin, and Hyochoong Bang}
%
\institute{Korea Advanced Institute of Science and Technology, Daejeon, Korea
\and
Dongguk University, Seoul, Korea  \\
\email{hcbang@kaist.ac.kr}\\
}

\maketitle              % typeset the title of the contribution

\begin{abstract}
This paper proposes a data-driven model predictive control for multirotor collision avoidance considering uncertainty and an unknown model from a payload. To address this challenge, sparse identification of nonlinear dynamics (SINDy) is used to obtain the governing equation of the multirotor system. The SINDy can discover the equations of target systems with low data, assuming that few functions have the dominant characteristic of the system. Model predictive control (MPC) is utilized to obtain accurate trajectory tracking performance by considering state and control input constraints. To avoid a collision during operation, MPC optimization problem is again formulated using inequality constraints about an obstacle. In simulation, SINDy can discover a governing equation of multirotor system including mass parameter uncertainty and aerodynamic effects. In addition, the simulation results show that the proposed method has the capability to avoid an obstacle and track the desired trajectory accurately. 

\end{abstract}
\section{Introduction} % 이동우, 김윤성, 한명(?)
% 1. 드론 배경 (김윤성)
Multirotor-type unmanned aerial vehicles (UAVs) are systems actively utilized in various fields beyond military applications, including civilian sectors. With increasing demand and interest in the multirotor, there has been ongoing research in various areas such as UAV hardware \cite{hardware}, control systems \cite{control1}, and autonomous flight \cite{autonomous1}. Research on multirotor hardware focuses on improving the performance of actuators and batteries, enabling a multirotor to transport heavy objects or fly for extended periods. The development of flight control hardware and autonomous flight technology allows drones to perform more complex tasks independently.
% 드론은 군사 용도 뿐만 아니라 민간 분야등 다양한 영역에서 활발히 활용되고 있는 항공기 시스템이다. 최근 드론의 높은 수요와 관심에 따라 드론의 하드웨어[1], 제어 시스템[2,3], 자율 비행[4,5] 등 다양한 연구가 지속되고 있다. 드론 하드웨어에 대한 연구는 액추에이터, 배터리 등 하드웨어의 성능을 향상시켜 드론이 무거운 물건을 수송하거나 장시간 비행이 가능하도록 한다. 비행 제어 시스템과 자율 비행의 개발은 드론이 더욱 복잡한 임무를 스스로 수행할 수 있도록 만들어 준다. 

% 2. 자율 비행의 중요성과 필요한 기술(?)  (김윤성)
Recently, there has been an increase in the use of multirotor for missions in remote exploration and emergency disaster situations \cite{uav_disaster1, uav_disaster2}. Autonomous flight technology enables multirotor to successfully complete missions by navigating complex environments and finding optimal routes without remote control. For autonomous flight, technologies such as path-planning, obstacle avoidance, and path-following are essential to guarantee a successful mission \cite{path_plan}. 
% 최근, 원격지 탐사 및 긴급 재난 상황에서 드론을 활용하여 임무를 수행하는 사례들이 늘어나고 있다 [6,7]. 자율 비행 기술은 원격 조종 없이 스스로 복잡한 환경에서 안정적인 비행과 최적의 경로를 통해서 미션을 성공적으로 수행할 수 있게 만든다. 특히, 드론의 자율 비행을 위해서는 비행 경로 선정, 장애물 회피, 비행 경로 추종과 같은 기술들이 필수적이다 [8].
% 3. 관련된 기법들 --> mpc 기반의 장애물 회피를 경로 추종 기법 (김윤성)

Path-following techniques for multirotor UAV are control methods that ensure that the system follows a given path accurately. These techniques can be categorized into geometric methods, vector fields, and model predictive control (MPC) \cite{survey:path_follow}. The geometric method involves following a virtual target point (VTP) on the given path, thereby guiding the multirotor along the path. Well-known examples of control-oriented algorithms in the geometric methods include the nonlinear guidance law (NLGL) \cite{nlgl} and carrot chasing method. The vector field method generates vectors at each point along the path, indicating the direction of movement to ensure that the multirotor follows the path \cite{vector_field}. Lastly, MPC is a control method that predicts future system behavior to calculate an optimal control input.
According to Lee et al. \cite{survey:adv_dis}, the geometric technique allows precise control by preventing deviation from the given path, but it has the disadvantage of being less adaptable to uncertainties, making it difficult to apply to quadrotors. The vector field method has the advantage of low computational cost, but as the environment becomes more complex, designing the vector field becomes increasingly difficult. Unlike other path-following methods, MPC has the advantage of performing optimal control even in nonlinear environments with constraints, which makes it applicable in various fields, such as autonomous driving, path planning \cite{mpc}, and collision avoidance. However, when a multirotor UAV is transporting supplies or rescuing personnel, its physical parameters, such as mass and moment of inertia, can dynamically change. In addition, the unknown aerodynamic model that occurs from an additional payload can affect the tracking performance in the MPC framework. To address the above problems, a novel data-driven MPC that updates a nominal model is proposed to enhance tracking performance when the system model is changed. 

Data-driven control is based on various data-driven modeling methods such as dynamic mode decomposition (DMD) \cite{SD1}, Koopman operator \cite{SD2}, and sparse identification of nonlinear dynamics (SIDNy) \cite{SD3}, Gaussian process. These methods can give a meaningful representation of system, unlike a deep neural network that performs as a black box. Among data-driven methods, SINDy requires low data to obtain a governing equation of system because this method employs the $L_1$ penalty to obtain a sparse model. Using this characteristic, SINDy can provide a meaningful representation of the system relative to conventional machine learning methods. Due to this advantage, SINDy has been implemented in a wide range of systems, including the battery \cite{SINDy-bat}, aircraft \cite{SD4,SINDy-air}, manipulator \cite{SD5}, and ship \cite{SD6}. In addition, SINDy-based model predictive control (SINDy-MPC) has also received a lot of attention in the control field because SINDy can give accurate model prediction information in the MPC framework \cite{SINDy-MPC}. In \cite{SINDy-MPC1}, SINDy-MPC is utilized for a high-performance trainer aircraft to obtain accurate attitude tracking performance. Bhadriraju et al. \cite{SINDy-MPC2} suggest a data-driven MPC for a chemical reactor using OASIS that is based on the SINDy method to update a model. Manzoor et al. \cite{SINDy-MPC3, SINDy-MPC4} propose an SINDy-based model predictive control framework from largely known physics-informed models for ducted-fan unmanned aerial vehicles to enhance tracking performance. Bhattacharya et al. \cite{SINDy-MPC5} introduce the application of SINDYc-based MPC to solve the closed-loop control problem of RoSE for achieving desired peristaltic waves. 

Recent studies have demonstrated the successful application of SINDy in constructing models for
multirotor systems in \cite{Quad1,Quad2,Quad3,Quad4}. Kelm et al. \cite{Quad1} propose a SINDy-based reconfiguration control method that updates the system model with a time window. In \cite{Quad2}, simplified quadrotor model is identified using SINDy. The SINDy-based LQR is suggested for a quadrotor to obtain precise control performance \cite{Quad3}. Lee et al. utilize the SINDy method for a quadrotor system, including the gyroscopic and aerodynamic effects to formulate a Thau observer.

Although SINDy has achieved success in the early stages of model construction, in practical applications, Refs. \cite{Quad1,Quad2,Quad4} do not consider the full nonlinear equation and Refs. \cite{Quad1,Quad2,Quad3} also cannot regard unknown models such as aerodynamic effects that are significant in forward flight. In addition, every research does not consider a model predictive control that can account for constraints of state and control input.

     Therefore, to address these limitations, we discover a full order nonlinear multirotor system by considering parameter uncertainties and unknown model effects. Moreover, we propose a data-driven model predictive control for multirotor using the obtained SINDy model. In addition, our research not only follows reference trajectories but also avoids obstacles. 

The paper makes the following contributions: 
\begin{enumerate}
    \item The data-driven modeling method can provide a meaningful representation for a multirotor system considering mass parameter uncertainty and aerodynamic effect. \\
    
    \item SINDy-MPC for collision avoidance is proposed to avoid the obstacle during flight mission. To our knowledge, this study is the first attempt for a collision avoidance of multirotor system using SINDy and MPC. 
\end{enumerate}

This article is organized as follows: Section \ref{sec2} briefly introduces multirotor modeling with payload. In Section \ref{sec3}, we demonstrate a concept of SINDy. Section \ref{sec4} illustrates how to obtain a data-driven model and formulate SINDy-MPC. Section \ref{sec5} shows the simulation results by comparing the tracking performance to nominal nonlinear model predictive control. Finally, Section \ref{sec6} recapitulates the main points of the paper. 

% 드론의 경로 추종 기법은 시스템이 주어진 경로를 따라 정확하게 이동하도록 제어하는 기술이다. 경로 추종 기법은 대표적으로 크게 geometric, vector field, Model Predictive Control (MPC) 등으로 나눌 수 있다 [11]. Geometric technique은 주어진 경로상에 존재하는 Virtual Target Point(VTP)를 쫒아가면서 경로를 따라가는 방법이다. Geometric technique에서 control-oriented algorithm의 가장 잘 알려진 예시로는 Non-Linear Guidance Law (NLGL) [12], carrot chasing 이 있다. Vector field는 주어진 경로를 따라 이동할 수 있도록 각 지점에서 이동 방향을 나타내는 벡터를 생성하는 방법으로, 생성된 벡터는 이동체가 경로를 따라가도록 유도한다 [13]. 마지막으로 MPC는 시스템의 미래 동작을 예측하여 최적의 제어 입력을 계산하는 제어 방법이다. 
% Lee et al. [14] 에 따르면, Geometric technique은 주어진 경로를 이탈하지 않도록하여 정밀한 제어가 가능하다는 장점이 있지만 disturbunce 및 불확실성에 잘 적응하지 못하는 특성이 있어 quadroter에 적용하기 힘들다는 단점이 있고, vector field는 computational cost가 낮다는 장점이 있지만, 환경이 복잡할수록 vector field의 설계가 어렵고 복잡하다는 단점이 있다. MPC는 다른 경로 추종 방식과 달리 제약 조건이 있는 비선형 환경에서도 최적제어를 수행할 수 있다는 장점이 있어 자율 주행 기술에 필요한 경로계획 [15], 충돌 회피 등 다양한 분야에 활용되고 있다. 특히, 드론이 물자를 운송하거나 구조자를 운반하는 경우에 무게, 크기 등 드론의 물리량이 유동적으로 변하게 된다. 이처럼 변화하는 환경에서 MPC와 같은 예측제어를 통한 자율 비행을 위한 기술이 요구되고 있다.
  
% 삭제
% 하지만 비선형 시스템의 경우 computational cost가 높아 실시간 시스템에 적용되기 힘들다는 단점이 있어, 이를 보완하는 기술이 요구되어진다.
% However, in nonlinear systems, the high computational cost makes real-time application challenging, necessitating the development of complementary technologies.

% 4. 데이터 기반 MPC (이동우)
% 5. 기존의 기법들의 한계점 (이동우)
% 6. 기여점 (이동우)
% 7. 논문의 개요 (이동우)

%한국어로 작성! 

\section{Modeling of Multirotor with Payload}
\label{sec2}
\begin{figure}[htb!]
\centering
\includegraphics[width=0.4\textwidth ]{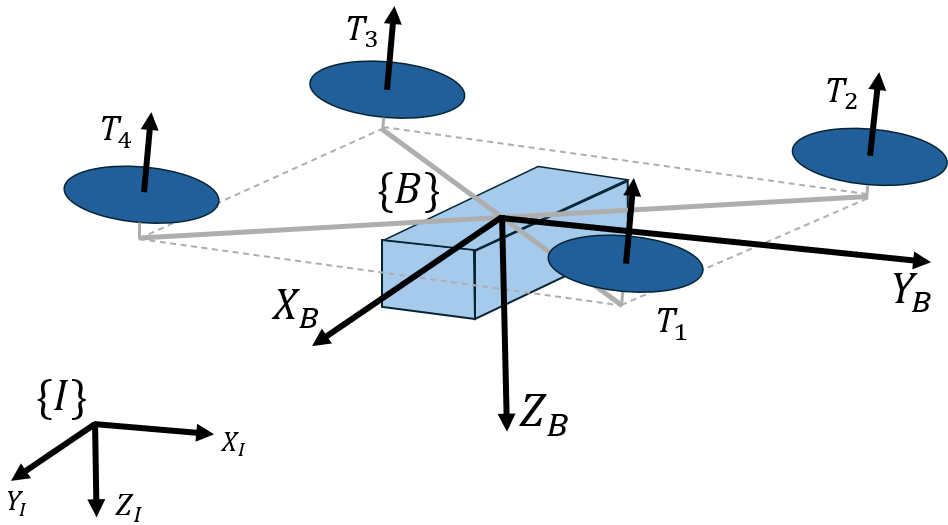}
\caption{Configuration of multirotor with payload.}
\label{fig1}
\end{figure}
The inertial frame $\{\bm{I}\}$ is defined as a North-East-Down (NED) frame with a fixed origin. The body-fixed frame $\{\bm{B}\}$ is defined with its axis $x_{B}$, $y_{B}$, and $z_{B}$ parallel to the forward, right, and descending directions from the center of gravity of the multirotor, respectively.  Figure \ref{fig1} illustrates the configuration of the 6-DOF multirotor dynamic model. 

%Inertial 좌표계 I는 고정된 원점을 가진 NED좌표계로 정의된다. Body-fixed 좌표계 B는 Quadrotor의 무게중심으로부터 전방, 오른쪽, 하강 방향에 각각 평행한 축 $x_{B}$, $y_{B}$, $z_{B}$로 정의된다. 그림1은 6 dof 쿼드롭터 동역학 모델의 configuration이다.
The position and Euler angles measured in the $\{\bm{I}\}$ frame are represented by vectors $ {\bm{\eta}} = \left[x, y, z \right]^{\top} \in {\mathbb{R}}^{3}$ and ${\bm{\Omega}} = \left[\phi, \theta, \psi \right]^{\top} \in {\mathbb{R}}^{3}$. The velocity in $\{\bm{I}\}$ frame is represented by vectors $ {\bm{\nu}} = \left[\dot{x}, \dot{y}, \dot{z} \right]^{\top} \in {\mathbb{R}}^{3}$. The angular rates measured in the $\{\bm{B}\}$ frame is represented by vectors $ {\bm{\omega}} = \left[p, q, r \right]^{\top} \in {\mathbb{R}}^{3}$.
The multirotor dynamics can be summarized as (\ref{eq:1}) - (\ref{eq:6}), where the following notations are used for brevity: $\sin (x) \to {s_x}{\rm{  }} ,\,\, \cos (x) \to {c_x}{\rm{ }},\,\, \tan (x) \to {t_x}{\rm{ }}$. 

%$I$ frame에서 픅정된 위치와 오일러각은 벡터 $ {\bm{\eta}} = \left[x, y, z \right]^T$ 와 ${\bm{\zeta}} = \left[\phi, \theta, \psi \right]^T$ 로 나타내며, $B$ frame에서 측정된 속도와 각속도는 벡터 $ {\bm{\nu}} = \left[u, v, w \right]^T$ 와 $ {\bm{\omega}} = \left[p, q, r \right]^T$으로 나타냅니다. 
%쿼드롭터 동역학을 정리하면 다음과 같이 정리할 수 있습니다.
\begin{equation}
\begin{split}
\dot{\bm{\eta}}  = \bm{\nu} ,
\end{split}
\label{eq:1}
\end{equation}
\begin{equation}
\renewcommand{\arraystretch}{0.6}
\dot{\bm{\Omega}}={\bf{R}}_{\omega}\bm{\omega},\\
\label{eq:2}
\end{equation}
\begin{equation}
\dot{\bm{\nu}} = \frac{1}{m_t} (\mathbf{R}_{B}^I \bm{{f}}_t + \bm{{f}}_a) -\bm{{g}}, \\
\label{eq:3}
\end{equation}
\begin{equation}
\dot{\boldsymbol{\omega}} = \bm{I}_t^{-1} \left( \boldsymbol{\tau}_t + {\bm{\tau}}_{a} - \boldsymbol{\omega} \times (\bm{I}_t \boldsymbol{\omega})\right),
\end{equation}
\label{eq:4}
\begin{equation}
\mathbf{R}_{B}^I = 
\begin{bmatrix}
 c_\theta c_\psi& s_\phi  s_\theta c_\psi - c_\phi s_\psi  &  c_\phi s_\theta  c_\psi + s_\phi s_\psi \\
c_\theta s_\psi & s_\phi s_\theta s_\psi + c_\phi c_\psi & c_\phi  s_\theta s_\psi - s_\phi c_\psi \\
-s_\theta & s_\phi c_\theta & c_\phi c_\theta 
\end{bmatrix},
\label{eq:5}
\end{equation}
\begin{equation}
\mathbf{R}_{\omega} = 
\begin{bmatrix}
1 & s_\phi t_\theta & c_\phi t_\theta \\
0 & c_\phi & -s_\phi \\
0 & \frac{s_\phi}{c_\theta}  & \frac{c_\phi}{c_\theta}
\end{bmatrix},
\label{eq:6}
\end{equation}
where $g$ and $m_t$ are the mass of multirotor with gravitational and payload acceleration, $\bm{g} =  \left[0, 0, g \right]^{\top}$ is the gravity vector, $\bm{f}_t = \left[0, 0, f_z \right]^{\top} \in {\mathbb{R}}^{3}$ and $\bm{{\tau}}_t = \left[L, M, N \right]^{\top} \in {\mathbb{R}}^{3}$ are the force and torques of the multirotor, $\bm{I}_t \in {\mathbb{R}}^{3 \times 3}$ is the inertia matrix of multirotor with payload, $\mathbf{R}_{B}^I$ is a transformation matrix from $\{\bm{B}\}$ frame to $\{\bm{I}\}$ frame, $\bm{R}_{\omega}$ are the rotation matrix between the angular velocity with respect to the $\{\bm{I}\}$ frame and angular rate in $\{\bm{B}\}$ frame.

%여기서 $m$, $g$는쿼드롭터의 질량과 중력가속도, $T$, ${{\tau}}$ 는 쿼드롭터의 추력과 토크이며,  $\tau_{\text{ext}}$ 는 공기역학적 외란 같은 외부 토크이며, $I _{v}$ 는 쿼드로터 전체의 관성행렬, $R _{BI}$ 와 $R _{q}$ 는 각각 위치와 오일러각에 대한 Rotation Matrix를 나타냅니다.

% For simplicity, the following notation has been used:
% %단, 표현을 간단히 하기 위해 다음과 같은 표기법이 사용되었습니다 : 
% $\sin (x) \to s_x \text{  }, \cos (x) \to c_x$.
%$I_v = diag([Ixx, Iyy, Izz])$
The thrust and moments generated by the four actuators can be expressed as follows:
%4개의 액추에이터가 생성하는 추력과 모멘트는 아래와 같이 나타낼 수 있으며,
\begin{equation}
\begin{bmatrix}
f_z \\
L \\
M \\
N
\end{bmatrix}
=
\begin{bmatrix}
-1 & -1 & -1 & -1 \\
-d_x & d_x & d_x & -d_x \\
d_y & d_y & -d_y & -d_y \\
c_{T} & -c_{T} & c_{T} & -c_T
\end{bmatrix}
\begin{bmatrix}
T_1 \\
T_2 \\
T_3 \\
T_4
\end{bmatrix},
\end{equation}
where $T_i$ is the thrust of $i^{\text{th}}$ actuator, $d_x$ and $d_y$ are the distances from the center of the drone to the actuators, and $c_{T}$ is the coefficient relating thrust to moments. 

Assuming non-diagonal moment of inertia (MoI) and center of gravity (CoG) 
 offset is zero, this payload only changes the mass properties such as mass and MoI. The mass and MoI of multirotor with payload can be defined as
 \begin{equation}
 \begin{split}
  m_{t} &= m_{m} + m_{p},\\
  \bm{I}_{t} &= \bm{I}_{m} + \bm{I}_{p},\\
 \end{split}
 \label{eq:Traslation_model13}
 \end{equation}
where $m_{m}$ and $m_{p}$ are the mass of multirotor and payload, $\bm{I}_{m} \in {\mathbb{R}}^{3 \times 3}$ and $\bm{I}_{p} \in {\mathbb{R}}^{3 \times 3}$ are the MoI of multirotor and payload. 

To account for aerodynamic forces and moments, we utilize a lumped aerodynamic model \cite{hanna} for the multirotor with payload as follows:
\begin{equation}
\begin{split}
    {\bm{f}}_{a} = \left[ {\begin{array}{*{20}{c}}
    {-K_{F,\dot{x}} \,\,\dot{x} }\\
    {-K_{F,\dot{y}} \,\,\dot{y} }\\
    {-K_{F,\dot{z}} \,\,\dot{z} }\\
    \end{array}} \right],{\bm{\tau}}_{a} = \left[ {\begin{array}{*{20}{c}}
    {-K_{M,p} \,\,\ p }\\
    {-K_{M,q} \,\,\ q }\\
    {-K_{M,r} \,\,\ r }\\
    \end{array}} \right],
\end{split}
\label{eq:Traslation_model5}
\end{equation}
\noindent where $K_{F,\dot{x}}, K_{F,\dot{y}}, K_{F,\dot{z}}$ are the aerodynamic force coefficients and $K_{M,p}, K_{M,q}, K_{M,r}$ are the aerodynamic moment coefficients.

\section{Sparse Identification of Nonlinear Dynamics}
\label{sec3}
Recently, although machine learning modeling methods have been widely used in various engineering fields, these approaches do not provide a meaningful representation of system and require a lot of data to learn the model. The SINDy method has the advantage of being able to acquire governing equations of target systems with low amount of data than conventional deep neural networks require. The main idea is to identify the sparse model from overall candidate functions, assuming that few functions have dominant characteristics of the system. To demonstrate this concept, we consider the continuous-time nonlinear system as
\begin{equation}
\begin{array}{l}
\dot{{\bf{x}}} = f(\mathbf{x}, \mathbf{u}),
\end{array}
\label{eq:sindy11}
\end{equation}
where ${\bf{x}} \in {\mathbb{R}}^m$ represents the state vector, ${\bf{u}} \in {\mathbb{R}}^n$ is the control input vector, and ${f}$ denotes the nonlinear system function.

\begin{figure}[htb!]
    \centering
    \includegraphics[width=0.7\textwidth]{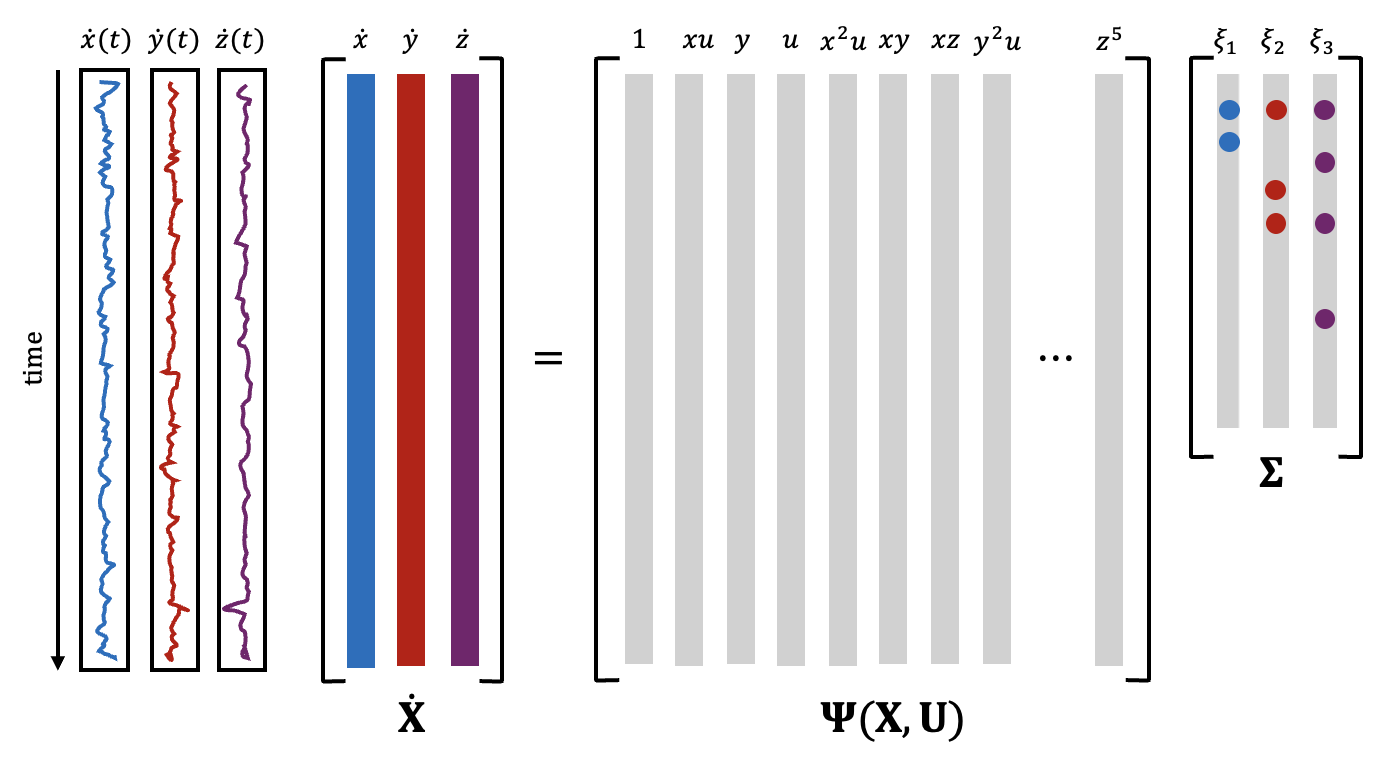}
    \caption[Concept of SINDy]{Concept of SINDy.}
    \label{fig:Schematic Controller1}
\end{figure}

To derive the sparse model, the system in Eq. (\ref{eq:sindy11}) can thus be written as follows: 
\begin{equation}
\begin{array}{l}
\dot{\textbf{X}} = \bf{\Psi}(\textbf{X},\textbf{U})\bf{\Sigma}, \\
\end{array}
\label{eq:sindy21}
\end{equation}
where ${\textbf{X}} \in {\mathbb{R}}^{w \times m}$ denotes the state snapshots of a system, ${\textbf{U}} \in {\mathbb{R}}^{w \times n}$ contains snapshots of control input, $\bf{\Sigma}$ is the coefficient of model, $\bf{\Psi}$ is the library of the candidate functions and $\dot{\textbf{X}} \in {\mathbb{R}}^{w \times m}$ represents the time derivative of $\textbf{X}$.

The snapshots of state \textbf{X} and snapshots of control input are defined as
\begin{equation}
\begin{split}
\textbf{X} &= \begin{bmatrix}
x(t_1) & x(t_2) & \cdots & x(t_w)
\end{bmatrix}^{\top}, \\
\textbf{U} &= \begin{bmatrix}
u(t_1) & u(t_2) & \cdots & u(t_w)
\end{bmatrix}^{\top} .\\
\end{split}
\end{equation}

In practice, $\dot{\textbf{X}}$ can be computed directly from the data in $\textbf{X}$. The numerical differential method such as simple forward
Euler finite-difference can be utilized to obtain these values. For noisy data, the total variation regularized derivative typically yields numerically robust results \cite{SINDy-MPC}. Alternatively, it can be computed using SINDy methods for discrete-time systems $\textbf{x}_{k+1} = \textbf{F}(x_{k})$, as in the DMD algorithm, and avoid derivatives entirely. As a result, time derivative $\dot{\textbf{X}}$ are sampled at several time ${t_1}$, ${t_2}$, $\cdots$, ${t_w}$ and arranged as follows:
\begin{equation}
\begin{split}
\dot{\textbf{X}} &= \begin{bmatrix}
\dot{x}(t_1) & \dot{x}(t_2) & \cdots & \dot{x}(t_w)
\end{bmatrix}^{\top}.\\
\end{split}
\label{sindy5}
\end{equation}

The library of candidate functions is constructed to include various types of nonlinear functions of the columns of $\textbf{X}$ and $\textbf{U}$. For example, $\bf{\Psi}$($\textbf{X}$, $\textbf{U}$) may consist of constant, polynomial, trigonometric, and exponential functions. To improve the model accuracy, physics-informed candidate functions can be used when we design the library function \cite{SINDy-MPC3, SINDy-MPC4}. We consider a library of candidate nonlinear functions $\bf{\Psi}$($\textbf{X}$, $\textbf{U}$) of the form as
\begin{equation} 
\begin{split}
    \bf{\Psi} (\textbf{X},\textbf{U})  = \left[ {\begin{array}{*{20}{c}}
    \mid & \mid &  \mid &  \mid & \mid  &  \mid & \mid  \\
       \textbf{1} & \textbf{X} & \textbf{U}&  \textbf{X}\otimes\textbf{X} & \cdots & sin(\bf{X})   & \cdots\\
       \mid & \mid &  \mid &  \mid & \mid &  \mid & \mid   \\
    \end{array}} \right],
\end{split}
\label{eq:sindy6}
\end{equation}
where $\textbf{x} \otimes \textbf{y}$ defines the vector of all product combinations of the components in $\textbf{x}$ and $\textbf{y}$. 

The regularization methods such as $L_1$ (Lasso), $L_2$ (Ridge), and elastic net regularizations are used to encourage sparsity in the solution. These methods can improve the robustness of identification for overdetermined data. The sparse regression problem is formulated as follows:
\begin{equation} 
\begin{split}
    {\bf{\Sigma}} = \arg \min {\left\| \dot{\textbf{X}} - {\bf{\Psi}}(\textbf{X},\textbf{U}){{\bf{\Sigma}}}    \right\|_2}^2 + \lambda {\left\| {\bf{\Sigma}} \right\|_1},
\end{split}
\label{eq:sindy423}
\end{equation}
where $\lambda$ is the regularization parameter promoting sparsity through an $L_1$ penalty.

\section{Sparse Identification of Nonlinear Dynamics-based Model Predictive Control for Collision Avoidance} % 쿠푸만 (이동우) + MPC (김윤성)
\begin{figure}[htb!]
    \centering
    \includegraphics[width=1.01\textwidth]{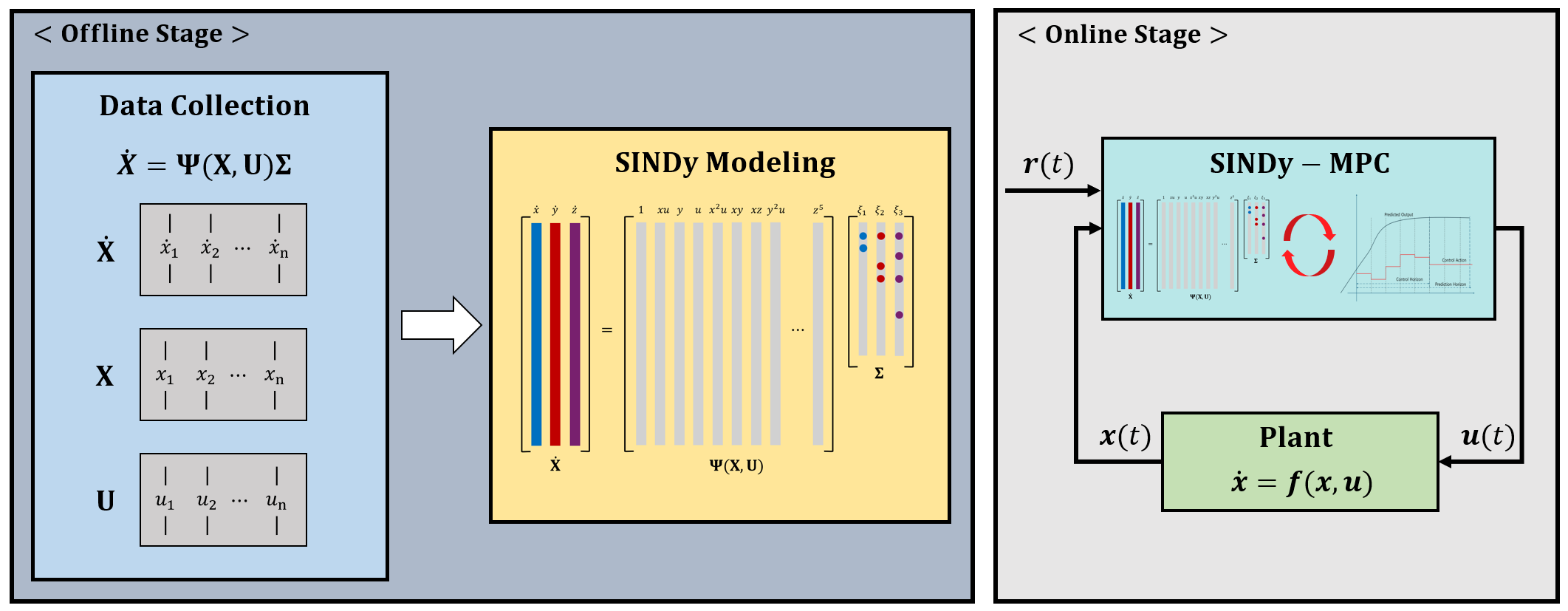}
    \caption[Schematic of SINDy-MPC.]{Schematic of SINDy-MPC.} 
    \label{fig:SchematicSINDYFTC}
\end{figure}

\label{sec4}
In this section, SINDy-MPC is introduced for trajectory tracking and collision avoidance to ensure autonomous flight operation. The concept of proposed method is demonstrated in Fig. \ref{fig:SchematicSINDYFTC}. The proposed control method is divided into an offline stage and an online stage. In the offline stage, firstly, we collect the state and control input data to discover a governing equation of multirotor. Secondly, we find the data-driven model using SINDy to implement the MPC. In the online stage, SINDy-MPC is utilized to obtain optimal control input by considering state constraints and obstacle inequalities. 
% The SINDy is employed to identify nonlinear and uncertain systems using only the input and output data sets. First, a PID controller is implemented in the quadrotor system to collect data. Second, we find a data-driven model using SINDy to apply MPC for autonomous flight. Finally, the proposed control method can avoid this when the static obstacle is near the desired trajectory. Therefore, this method can allow us to use optimal control in a quadrotor system. 
\subsection{Data Collection for SINDy}
\begin{figure}[htb!]
    \centering
    \includegraphics[width=0.8\textwidth]{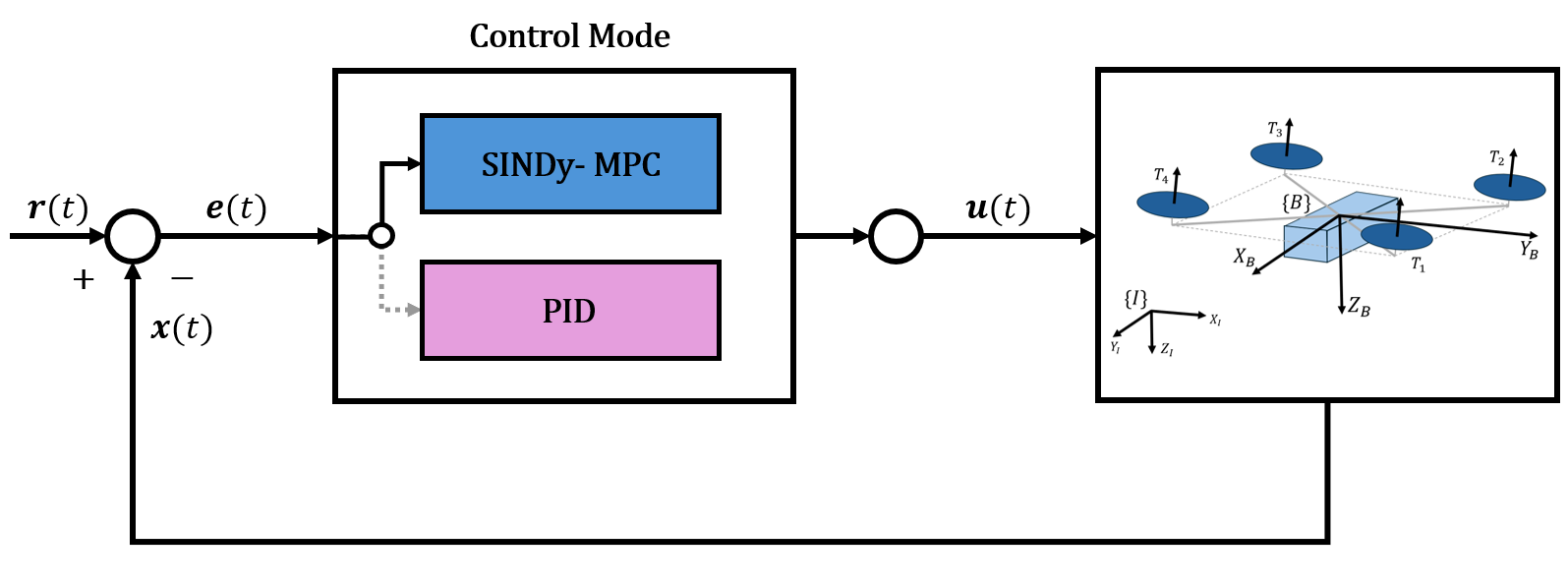}
    \caption[Schematic of control mode to collect data.]{Schematic of control mode to collect data.} 
    \label{fig:SchematicSINDYFTC1}
\end{figure}

To obtain an accurate model, we should obtain rich data to discover a target system. The basic strategy to collect data is to use random control input to avoid an imbalance. For example, if our target system is an automobile, we can give arbitrary wheel angle and lateral acceleration as control input while not considering the stability of system. However, aerial system cannot possibly use random input to avoid data bias. Therefore, in this paper, we utilize a classic PID controller that is usually used in the industrial field but cannot guarantee precise control performance to gather enough data. Moreover, if it is possible to operate a multirotor by a skilled pilot, this data-collecting method is also a suitable way for the aerial system because we can accumulate various data during flight operation. Figure \ref{fig:SchematicSINDYFTC1} shows a overall process about data-collecting.
\newpage
\subsection{Modeling of Multirotor using SINDy}
When we formulate a nominal model for MPC, we should need a dynamics of multirotor system because a kinematics is known in advance. Therefore, we should find a translational and rotational dynamics of system using a SINDy method. Firstly, we will consider the translational motion of multirotor, designing the state and control input as follows: 
\begin{equation} 
\begin{split}
\begin{array}{l}
 {\bf{x}}_{tr} = {\left[ {\begin{array}{*{20}{c}}
\dot{x}, &  \dot{y}, & \dot{z}\\
\end{array}} \right]^{\top}}, \\ 
 {\bf{u}}_{tr} = {\left[ {\begin{array}{*{20}{c}}
 \mathbf{R}_{B}^I (1,3) f_z, & \mathbf{R}_{B}^I (2,3) f_z, & \mathbf{R}_{B}^I (3,3) f_z\\
\end{array}} \right]^{\top}},\\ 
 \end{array}
\end{split}
\label{eq:sindymracinput1}
\end{equation}
where ${\bf{x}}_{tr} \in {\mathbb{R}}^{3}$ is the translational state vector, ${\bf{u}}_{tr} \in {\mathbb{R}}^{3}$ is the translational control input vector, $\mathbf{R}_{B}^I (i,j)$ is the rotational matrix element in $i^{th}$ row and $j^{th}$ column, and subscript $tr$ represents the translational term. 

A candidate function should be defined to apply SINDy. The library of candidate functions consists of prior knowledge of system and polynomial function. The prior function is designed by considering a kinematics and dynamics of multirotor in Eqs. (\ref{eq:1}) - (\ref{eq:6}).
\begin{equation} 
\begin{split}
{\bf{\Psi}}_{tr} (\textbf{X}_{tr}, \bf{\Omega}) = \left[ {\begin{array}{*{20}{c}}
       {\bf{\Psi}}_{tr,prior} (\textbf{X}_{tr},\textbf{U}_{tr}, {\bf{\Omega}}) \,\,\,\, {\bf{\Psi}}_{tr,poly} (\textbf{X}_{tr})
       \end{array}} \right], \\
\end{split}
\label{eq:sindy44}
\end{equation}
where
\begin{equation} 
\begin{split}
    {\bf{\Psi}}_{tr,prior} ({\bf{x}}_{tr},{\bf{u}}_{tr}, \bm{\Omega}) &= \left[ {\begin{array}{*{20}{c}}
       \mathbf{R}_{B}^I (1,3) f_z, & \mathbf{R}_{B}^I (2,3) f_z, & \mathbf{R}_{B}^I (3,3) f_z 
       \end{array}} \right], \\
    {\bf{\Psi}}_{tr,poly} ({\bf{x}}_{tr}) &= \left[ {\begin{array}{*{20}{c}}
       1 & \dot{x} & \cdots & \dot{x}^2 & \cdots & \dot{z}^d
    \end{array}} \right],
\end{split}
\label{eq:sindy45}
\end{equation}
where $d$ is the order of polynomial function, ${\bf{\Psi}}_{tr,prior}$ is the prior library function about translational motion, ${\bf{\Psi}}_{tr,poly}$ is the polynomial library function about translational motion, ${\textbf{X}}_{tr} \in {\mathbb{R}}^{w \times 3}$ denotes the state snapshots of a translational motion, ${\textbf{U}}_{tr} \in {\mathbb{R}}^{w \times 3}$ contains snapshots of control input of a translational motion, ${\bf{\Omega}} \in {\mathbb{R}}^{w \times 3}$ contains snapshots of Euler angles and $\dot{\textbf{X}}_{tr} \in {\mathbb{R}}^{w \times 3}$ represents the time derivative of $\textbf{X}_{tr}$.

Secondly, we will consider the rotational motion of multirotor, designing the state and control input as follows: 
\begin{equation} 
\begin{split}
\begin{array}{l}
 {\bf{x}}_{ro} = {\left[ {\begin{array}{*{20}{c}}
  p, & q, & r  \\
\end{array}} \right]^\mathrm{\top}}, \\ 
 {\bf{u}}_{ro} = {\left[ {\begin{array}{*{20}{c}}
   L, & M, & N \\
\end{array}} \right]^\mathrm{\top}}, \\ 
 \end{array}
\end{split}
\label{eq:sindymracinput6}
\end{equation}
where ${\bf{x}}_{ro} \in {\mathbb{R}}^{3}$ is the rotational state vector, ${\bf{u}}_{ro} \in {\mathbb{R}}^{3}$ is the rotational control input vector,  and subscript $ro$ represents the rotational term. 

The candidate function about rotational function is also defined using prior knowledge of system to enhance the accuracy of data-driven model as:
\begin{equation} 
\begin{split}
{\bf{\Psi}}_{ro} (\textbf{X}_{ro},\textbf{U}_{ro}) = \left[ {\begin{array}{*{20}{c}}
       {\bf{\Psi}}_{ro,prior} (\textbf{X}_{ro},\textbf{U}_{ro}) \,\,\,\, {\bf{\Psi}}_{ro,poly} (\textbf{X}_{ro})
       \end{array}} \right], \\
\end{split}
\label{eq:sindy445}
\end{equation}
where
\begin{equation} 
\begin{split}
    {\bf{\Psi}}_{prior} ({\bf{x}}_{ro},{\bf{u}}_{ro}) &= \left[ {\begin{array}{*{20}{c}}
      L&M&N&pq&qr&pr
       \end{array}} \right], \\
    {\bf{\Psi}}_{poly} ({\bf{x}}_{ro}) &= \left[ {\begin{array}{*{20}{c}}
       1 & p & \cdots & p ^2 & \cdots & r^w
    \end{array}} \right],
\end{split}
\label{eq:sindy454}
\end{equation}
where $w$ is the order of polynomial function about rotational motion, ${\bf{\Psi}}_{ro,prior}$ is the prior library function about rotational motion, ${\bf{\Psi}}_{ro,poly}$ is the polynomial library function about rotational motion, ${\textbf{X}}_{ro} \in {\mathbb{R}}^{w \times 3}$ denotes the state snapshots of a rotational motion, ${\textbf{U}}_{ro} \in {\mathbb{R}}^{w \times 3}$ contains snapshots of control input of a rotational motion, and $\dot{\textbf{X}}_{ro} \in {\mathbb{R}}^{w \times 3}$ represents the time derivative of $\textbf{X}_{ro}$.

Consequently, the sequential thresholded least squares (STLS) method is employed to obtain the model parameter, yielding a sparse model of the multitorotor system as described using Eqs. (\ref{eq:sindy4232}) and  (\ref{eq:sindy424}). 
\begin{equation} 
\begin{split}
    {\bf{\Sigma}}_{tr} = \arg \min {\left\| \dot{\textbf{X}}_{tr} - {\bf{\Psi}}_{tr}(\textbf{X}_{tr},\textbf{U}_{tr}, {\bf{\Omega}}){{\bf{\Sigma}}}_{tr}    \right\|_2}^2 + \lambda_{tr} {\left\| {\bf{\Sigma}}_{tr} \right\|_1},
\end{split}
\label{eq:sindy4232}
\end{equation}
\begin{equation} 
\begin{split}
    {\bf{\Sigma}}_{ro} = \arg \min {\left\| \dot{\textbf{X}}_{ro} - {\bf{\Psi}}_{ro}(\textbf{X}_{ro},\textbf{U}_{ro}){{\bf{\Sigma}}}_{ro}    \right\|_2}^2 + \lambda_{ro} {\left\| {\bf{\Sigma}}_{ro} \right\|_1},
\end{split}
\label{eq:sindy424}
\end{equation}
where ${\bf{\Sigma}}_{tr}$ is the coefficient of translational model, ${\bf{\Sigma}}_{ro}$ is the coefficient of rotational model, $\lambda_{tr}$ and $\lambda_{ro}$ are the regularization parameter promoting sparsity through an $L_1$ penalty.

Therefore, we can formulate the approximated full-order nonlinear multirotor system using SINDy as follows:
\begin{equation}
\begin{split}
\dot{\bf{x}} = \left[ {\begin{array}{*{20}{c}}
    \dot{\bm{\eta}} \\
    \dot{\bm{\nu} } \\
    \dot{\bm{\Omega}} \\
    \dot{\bm{\omega}}
    \end{array}} \right] \approx \left[ {\begin{array}{*{20}{c}}
    \bm{\nu} \\
    {\bf{\Psi}}_{tr} {\bf{\Sigma}}_{tr} \\
    {\bf{R}}_{\omega}\bm{\omega} \\
    {\bf{\Psi}}_{ro} {\bf{\Sigma}}_{ro}
    \end{array}} \right] = \hat{f}(\mathbf{x}, \mathbf{u}).
\end{split}
\label{eq:111}
\end{equation}

\subsection{Model Predictive Control using SINDy} % 박상혁
\begin{figure}[htb!]
\begin{center}
\includegraphics[width=0.7\textwidth]{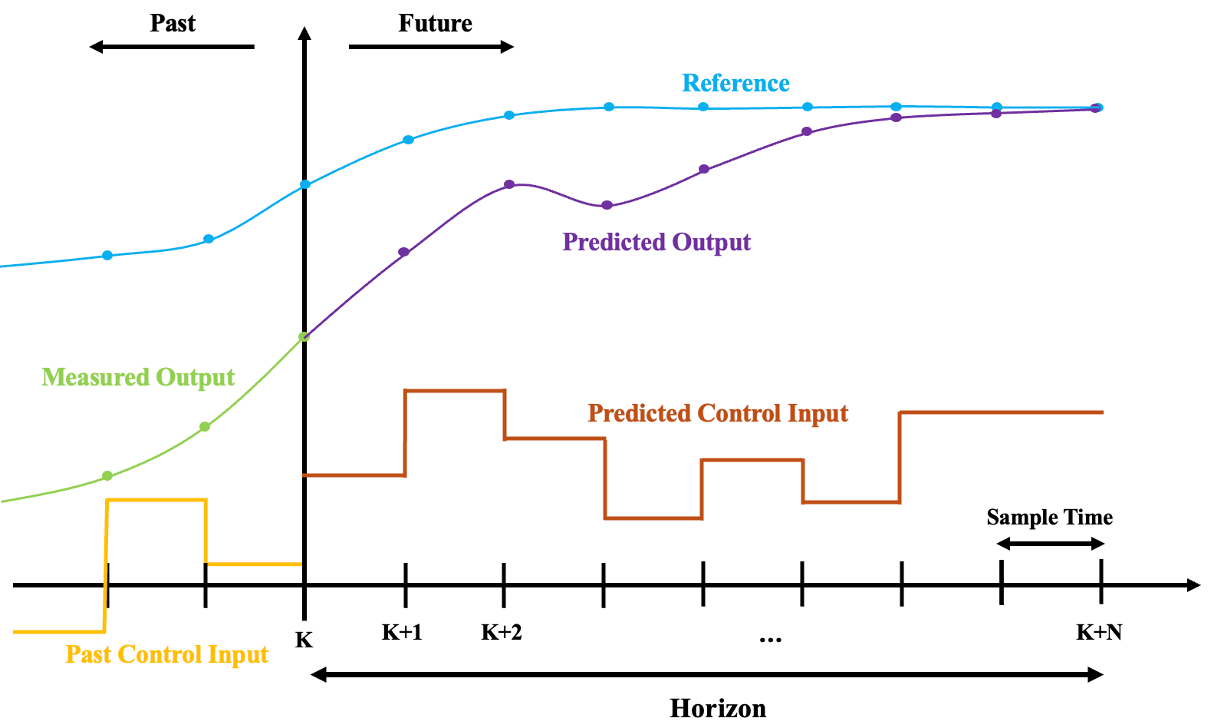}
\caption{\label{MPC} Concept of model predictive control.}
\end{center}
\end{figure}

 MPC is an optimal control strategy that predicts states and control inputs at the next time-step to minimize a cost function while considering some constraints.  By leveraging features of MPC, the optimal control input that guides the future state towards the desired trajectory is calculated by solving an optimization problem. The concept of MPC is illustrated in Fig. \ref{MPC}. For optimal control input, having an accurate system model is important since the predicted state affects MPC performance. Therefore, a data-driven modeling method is used to discover the system model in advance. Using the advantage of MPC that can handle constraints of state and control input, the multirotor can avoid the obstacle by adding the inequality constraint that prohibits the multirotor position near to the obstacle region. It is necessary to solve an optimization problem subject to equality and inequality constraints to achieve the optimal control input. In the MPC optimization problem, cost function consists of quadratic form of state error and control input. Therefore, SINDy-MPC problem and constraints are defined as
\begin{equation}
\begin{split}
    \displaystyle \min_{u_1,\cdots u_{N}} &\sum_{k=1}^{N}((r_i - y_i))^{\top} \bm{Q} (r_i - y_i)+ u_i^{\top} \bm{R} u_i),\\
    \text{s.t.} \,\, &\dot{\bm{x}} = \hat{f}(\bm{x},\bm{u}), \\
    \,\,\,\,\,\,\,\,\,\, & {u}_{min} \leq \bm{u} \leq {u}_{max}, \\
    \,\,\,\,\,\,\,\,\,\, & \bm{x}_{1} = \bm{x}_{init}, \\
    \,\,\,\,\,\,\,\,\,\, & \sqrt{ (x_{ob} - x)^2 + (y_{ob} - y)^2 + (z_{ob} - z)^2} \geq D_{min}, 
\label{eq.1}
\end{split}
\end{equation}
where $\bm{r}$ is the reference trajectory, $\hat{f}$ is the data-driven multirotor model from SINDy, $u_{min}$ and  $u_{max}$ are the minimum and maximum control input range, $\bm{x}_{init}$ is the initial condition of state, $\bm{Q}$ is the final weight matrix of error vector, $\bm{R}$ is the final weight matrix of control input vector, $N$ is the moving horizon step, $x_{ob},y_{ob}$ and $z_{ob}$ are the obstacle position, and $D_{min}$ is the minimum distance to guarantee the safety of the multirotor. To formulate the MPC, we use an open-source ACADO \cite{ACADO}. 

\section{Simulation Results} % 이동우
\label{sec5}
This section presents a simulation environment and validation results. In addition, we compare the trajectory tracking performance with nominal MPC with parameter uncertainty and an unknown model to validate the proposed control method.  
\subsection{Simulation Environment}
\begin{figure}[htb!]
    \centering
    \includegraphics[clip, width=8.5cm]{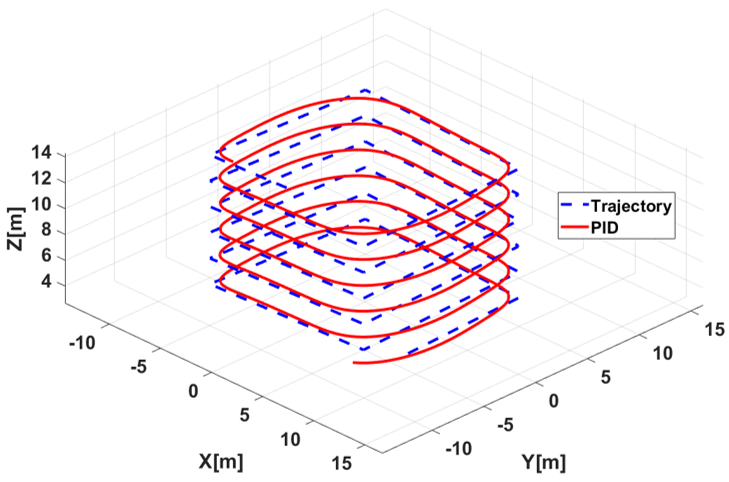}
    \caption[Schematic of K-FTC]{Data collection using PID controller.} 
    \label{fig:PID}
\end{figure}

Datasets for the data-driven model are acquired using the MATLAB simulator with a time step of $T_s$ = 0.002 s. To collect enough data, we define a simulation time of 100 s and a rectangular trajectory is used to achieve various movements using a PID controller \cite{PID}. Figure \ref{fig:PID} shows the result of trajectory following using PID controller for position tracking. The actual model parameters can be summarized in Table \ref{T1}.
\begin{table}[htb!]
\centering
\caption{Multirotor model parameters. }
\label{T1}
\begin{tabular}{|c|c|c|c|c|c|}
  \hline
Parameters & Values  & Units & Parameters & Values  & Units\\
  \hline
$I_{xx}$ & 0.0281 & $\textnormal{kg} \cdot \textnormal{m}^2$ & $d_x, d_y$    & 0.165           & $\textnormal{m}$\\
  \hline
$I_{yy}$ &  0.0286 & $\textnormal{kg} \cdot \textnormal{m}^{2}$ & $c_{T}$    & $0.0135$ & $\textnormal{m} $\\
  \hline
$I_{zz}$ &  0.0551 &$ \textnormal{kg} \cdot \textnormal{m}^{2}$ & $m$    & 1.3          & $\textnormal{kg}$\\
  \hline
  $K_{M,p},K_{M,q},K_{M,r}$    & $0.001$& $ \textnormal{N} / \textnormal{rad} / \textnormal{s} $ & $K_{F,\dot{x}},K_{F,\dot{y}},K_{F,\dot{z}}$    & $1.0$& $ \textnormal{N} / \textnormal{m} / \textnormal{s} $\\
  \hline
\end{tabular}
\end{table}
\subsection{Modeling and Validation}
\begin{figure}[htb!]%
\centering
\includegraphics[width=60mm,clip]{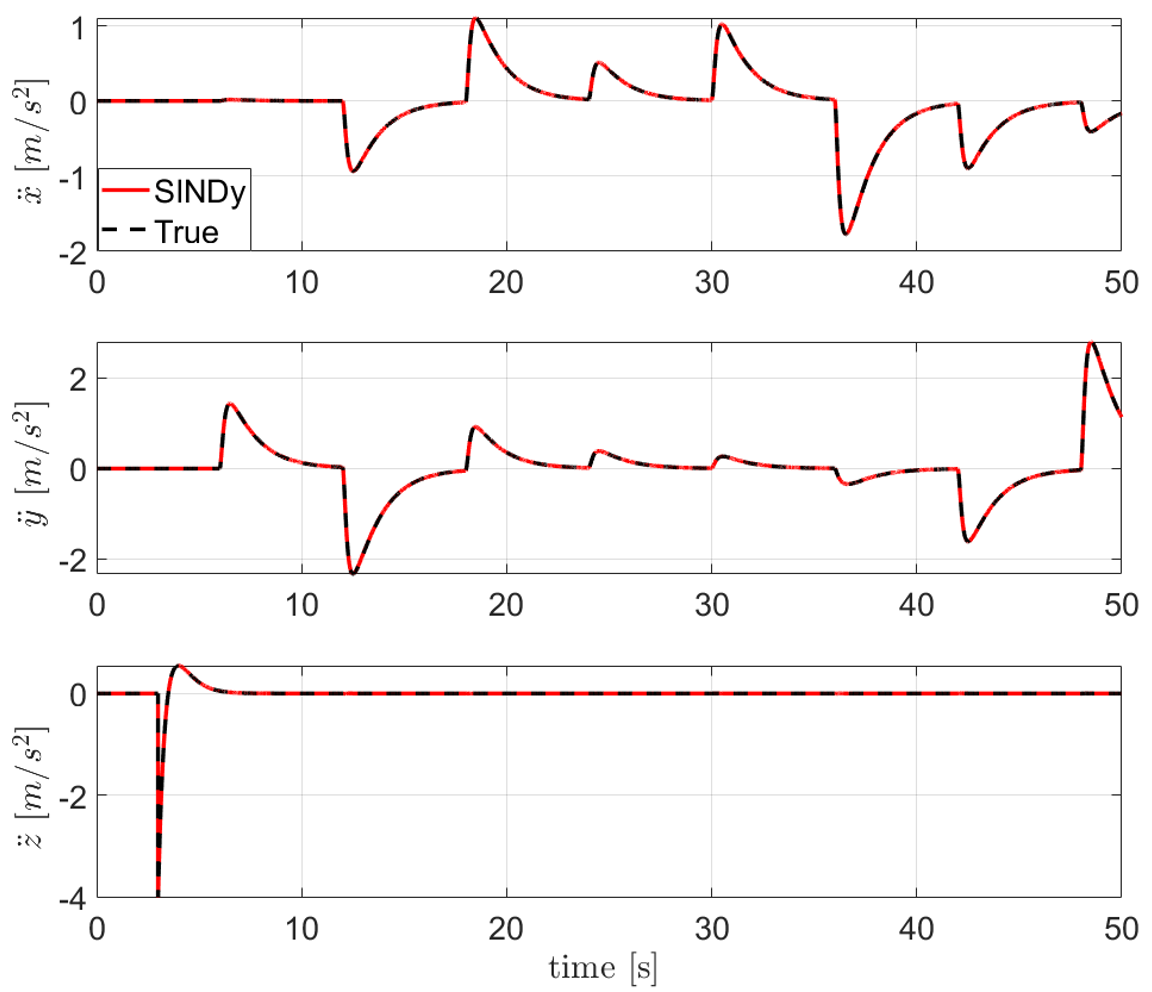}
\includegraphics[width=60mm,clip]{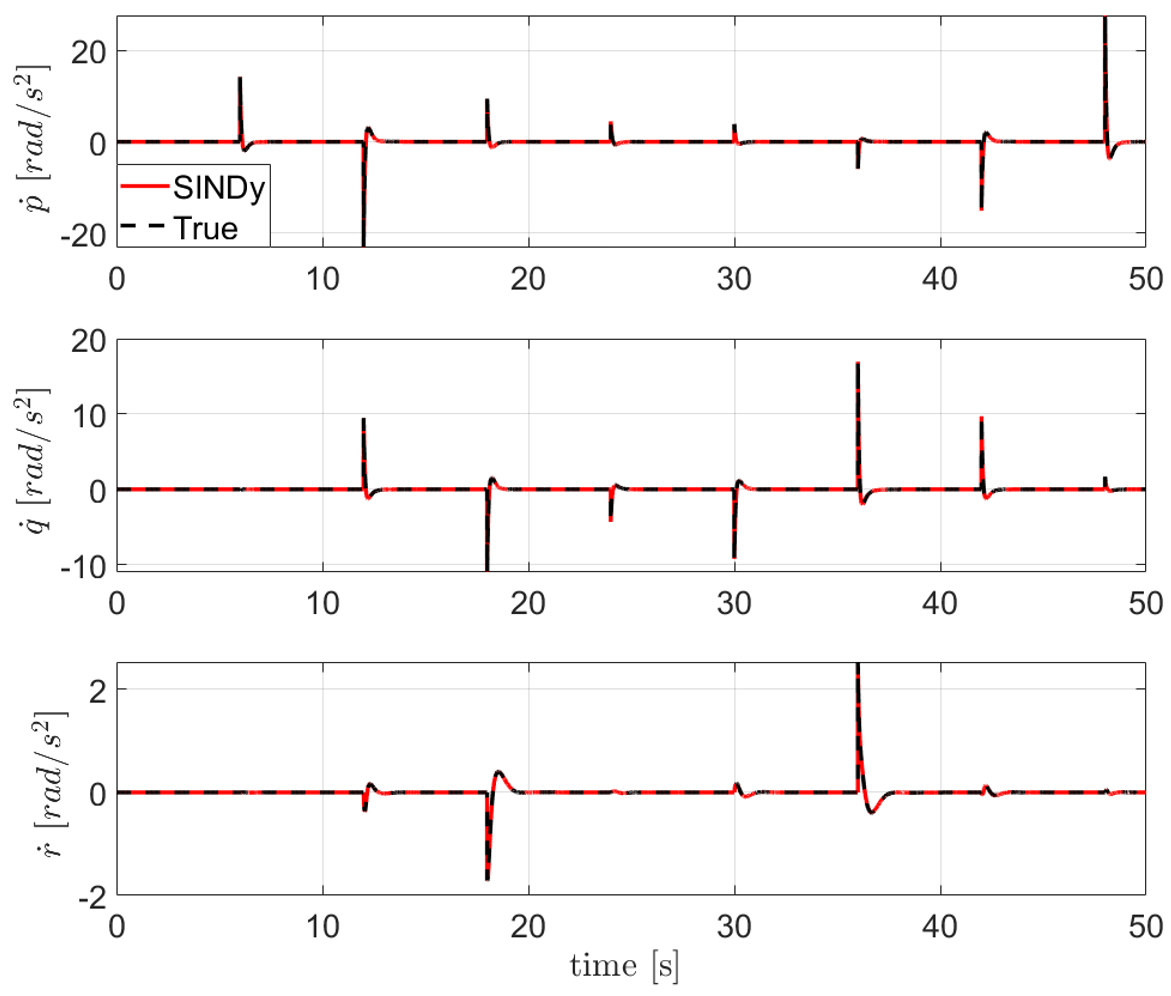}
  \caption{Validation of translational and rotational acceleration using SINDy.}
  \label{SINDy_val}%
\end{figure}

  Figure \ref{SINDy_val} shows the performance of translational and rotational acceleration prediction using the SINDy method. The data-driven model shows an accurate prediction performance compared to the one-step prediction. In translational motion, the obtained parameters consist of a thrust term and aerodynamic effect. Within the rotational model, the system model is composed of the following parameters: a coupling term, an aerodynamic term, and the control input. The true parameters $\Sigma_{(\cdot)}$ and identified parameters $\hat{\Sigma}_{(\cdot)}$ are shown in Tables \ref{T3} and \ref{T4}. Most identified parameters have an error within about $3\%$ of the true parameter.  However, the estimated aerodynamic coefficient on the yaw axis has about $33\%$ error because we do not give enough command to identify this yaw damping effect during data collection. Unlike this aerodynamic parameter, the overall data-driven models are obtained accurately to utilize this model as a nominal model in the MPC framework. 
  % To apply an MPC, we need an accurate model in multiple-step prediction so that the root mean square error (RMSE) is summarized along with the multiple-step in Table \ref{T2}. The overall performance is suitable for predicting the future state because RMSE of state is below 3$\%$ at N = 100. This result demonstrates that the moving horizon can be used up to 100. 
\begin{table}[htb!]
\centering
\caption{Result of translational model identification.}
\label{T3}
\label{table:faultIdentification}
\begin{tabular}{|c|c|c|c|c|c|c|} 
  \hline
      & $\Sigma_{\dot{x}}$ & $\hat{\Sigma}_{\dot{x}}$ & $\Sigma_{\dot{y}}$ & $\hat{\Sigma}_{\dot{y}}$ & $\Sigma_{\dot{z}}$ & $\hat{\Sigma}_{\dot{z}}$ \\
   \hline
    $R(1,3) \cdot F_{B}$ &  0.7692 & \textbf{0.7707} & {0} & {0}  & 0 & 0   \\
    \hline
    $R(2,3) \cdot F_{B}$ & {0} &  0 &  0.7692 & \textbf{0.7704}  & 0  & 0 \\
    \hline
    $R(3,3) \cdot F_{B}$ & {0} &  0 & 0 & 0 & 0.7692  & \textbf{0.7721} \\
    \hline
    $1$ & 0 &  0 & 0 & 0 & 9.807  & \textbf{9.808} \\
    \hline
    $\dot{x}$ &-0.7692 &\textbf{-0.7705} & 0 & 0 & 0  & 0 \\
   \hline
    $\dot{y}$ & 0 &  0 & -0.7692 & \textbf{-0.7702} & 0  & 0 \\
    \hline
    $\dot{z}$ & 0 &  0 & 0 & 0 & -0.7692  & \textbf{-0.7701} \\
    \hline
    $\dot{x}^2$ & {0} &  0 & 0 & 0 & 0  & 0 \\
   \hline
    $\dot{y}^2$ & {0} &  0 & 0 & 0 & 0  & 0 \\
    \hline
    $\dot{z}^2$ & {0} &  0 & 0 & 0 & 0  & 0 \\
    \hline
    $\dot{x} \dot{y}$  & {0} &  0 & 0 & 0 & 0  & 0  \\
   \hline
    $\dot{y} \dot{z}$  & {0} &  0 & 0 & 0 & 0  & 0  \\
    \hline
    $\dot{x} \dot{z}$  & {0} &  0 & 0 & 0 & 0  & 0  \\
    \hline
\end{tabular}
\end{table}

\begin{table}[hbt!]
\centering
\caption{Result of rotational model identification.}
\label{T4}
\label{table:faultIdentification1}
\begin{tabular}{|c|c|c|c|c|c|c|} 
  \hline
      & $\Sigma_{p}$ & $\hat{\Sigma}_p$ & $\Sigma_{q}$ & $\hat{\Sigma}_q$ & $\Sigma_{r}$ & $\hat{\Sigma}_r$ \\
   \hline
    $L$ & 32.258 &  \textbf{32.248} & 0 & 0 & 0  & 0 \\
    \hline
    $M$ & 0 &  0 &  26.316 & \textbf{26.316} & 0  & 0 \\
    \hline
    $N$ & 0 &  0 & 0 & 0 & 15.873  & \textbf{15.879} \\
    \hline
    $pq$ & 0 &  0 & 0 & 0 & -0.1111  & \textbf{-0.1114} \\
   \hline
    $qr$ & 0.8065 &  \textbf{0.7991} & 0 & 0 & 0  & 0 \\
     \hline
    $pr$ & 0 &  0 & 0.8421 & \textbf{0.8643} & 0  & 0 \\
    \hline
    $p\Omega$ &0 &  0& -0.0179 & \textbf{-0.0175} & 0 & 0 \\
    \hline
    $q\Omega$ & -0.0219 &  \textbf{-0.0217} & 0 & 0 & 0  & 0 \\
    \hline
    $1$ & 0 &  0 & 0 & 0 & 0  & 0 \\
    \hline
    $p$ &  -0.0323 &  \textbf{-0.0329} & 0 & 0 & 0  & 0 \\
   \hline
    $q$ & 0 &  0 & -0.0263 & \textbf{-0.0257} & 0  & 0 \\
    \hline
    $r$ &  0 &  0 & 0 & 0 & -0.0159  & \textbf{-0.0233} \\
    \hline
    $p^2$ & {0} &  0 & 0 & 0 & 0  & 0 \\
   \hline
    $q^2$ & {0} &  0 & 0 & 0 & 0  & 0 \\
    \hline
    $r^2$ & {0} &  0 & 0 & 0 & 0  & 0 \\
    \hline
\end{tabular}
\end{table}
\subsection{Trajectory Tracking with Obstacle}
\begin{figure}[htb!]
    \centering
    \includegraphics[clip, width=10cm]{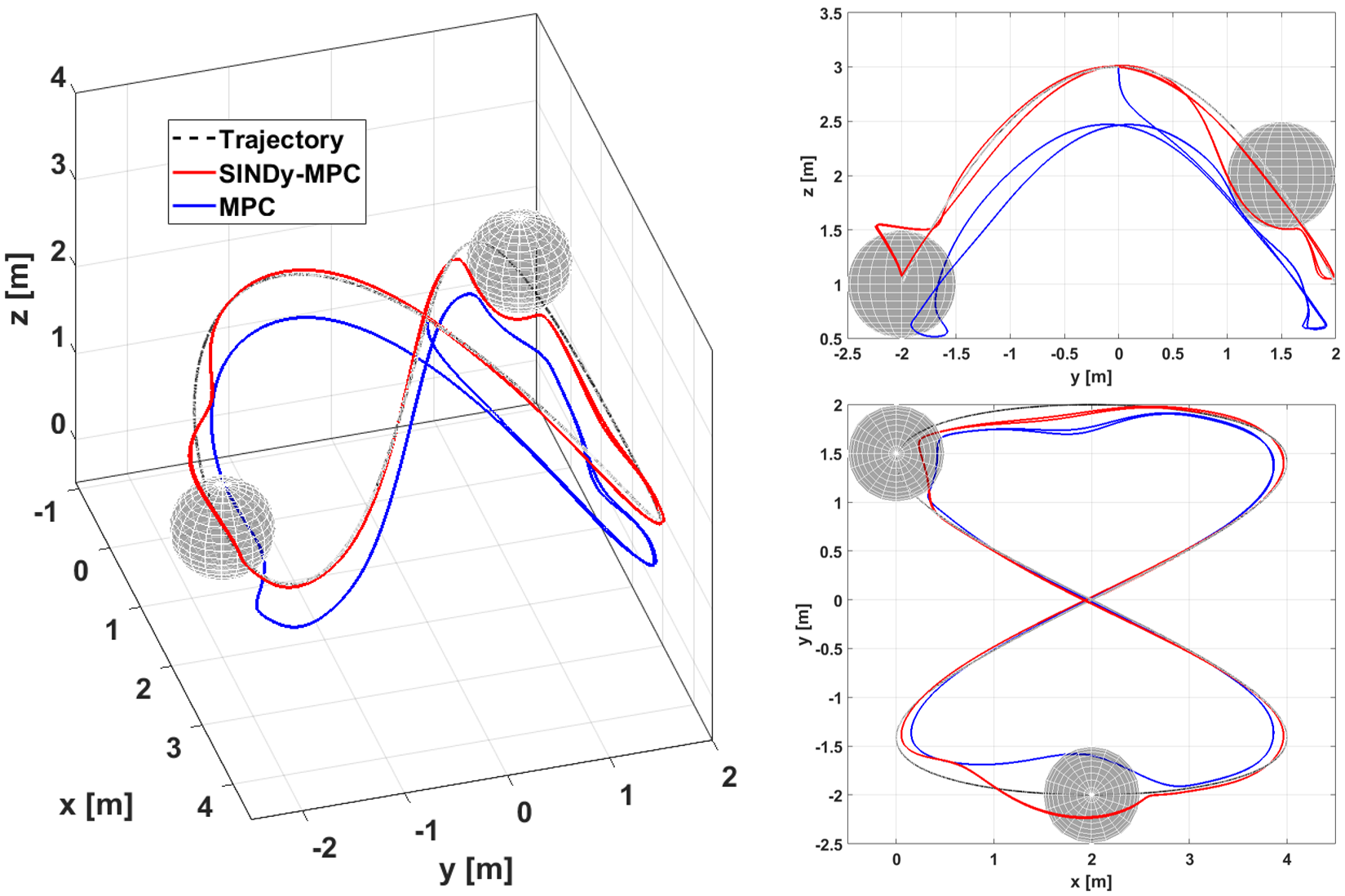}
    \caption{Trajectory tracking performance of SINDy-MPC and MPC.} 
    \label{SINDy-MPC1}
\end{figure}

\begin{figure}[htb!]
    \centering
    \includegraphics[clip, width=11.5cm]{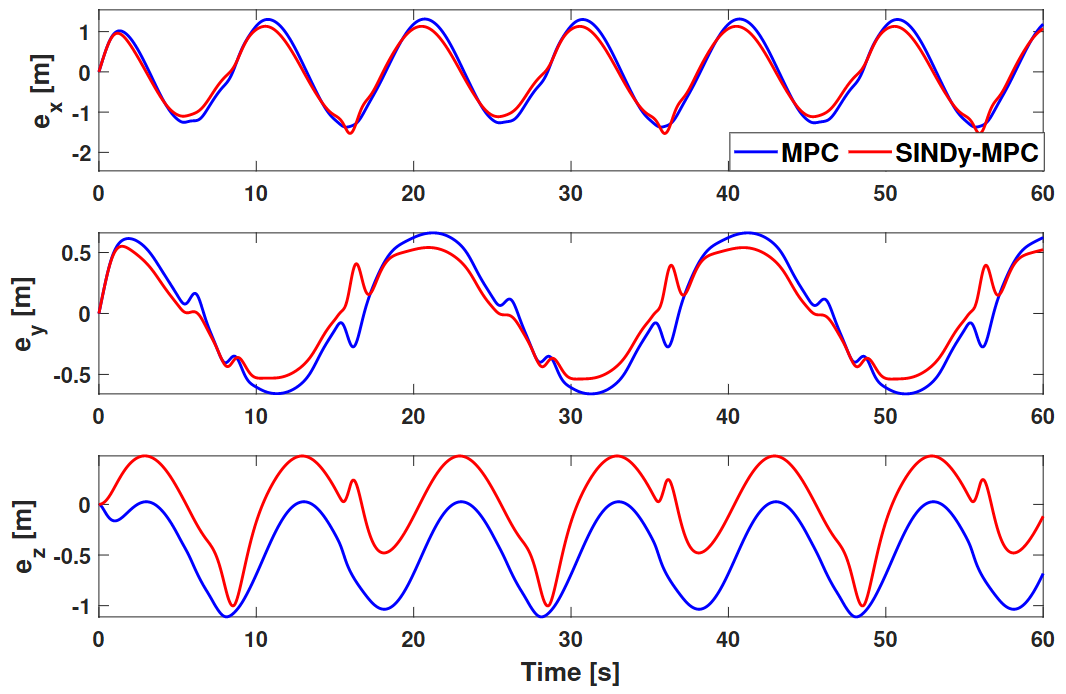}
    \caption[Schematic of K-FTC]{Trajectory tracking error.} 
    \label{SINDy-MPC2}
\end{figure}

\begin{figure}[htb!]%
\centering
\includegraphics[width=60mm,clip]{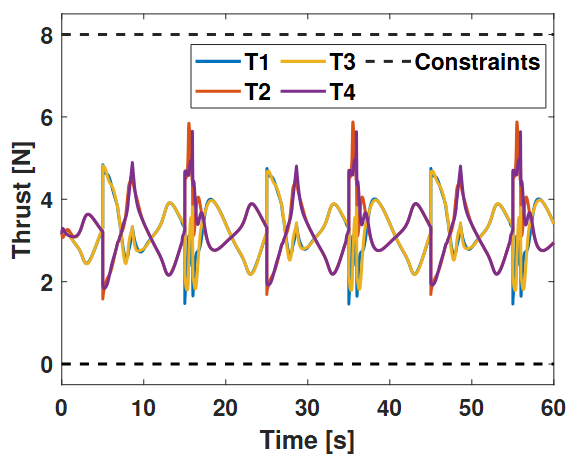}
\includegraphics[width=60mm,clip]{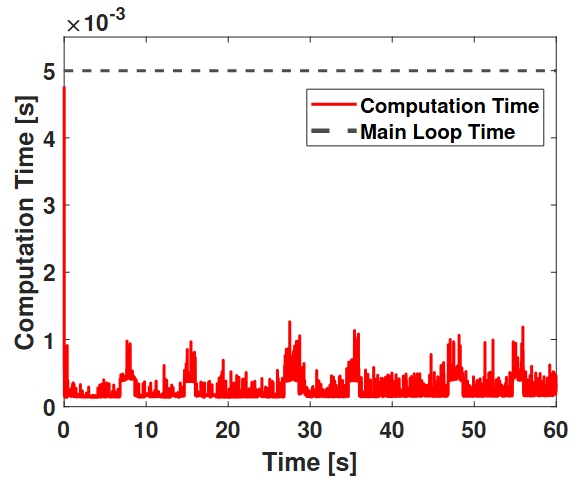}
  \caption{Thrust result (left) and computation time (right).}
  \label{SINDy-Time}%
\end{figure}

To confirm the capability of our method, we use the MPC that has an uncertainty of the mass parameter $20 \%$ and an unknown aerodynamic model due to a payload. Figure \ref{SINDy-MPC1} shows the position tracking performance of SINDy-MPC and MPC. Due to the uncertainty and aerodynamic model, the MPC cannot follow the trajectory, especially the altitude cannot compensate for the mass uncertainty effect. The proposed method can accurately follow the trajectory without the obstacle collision. Figure \ref{SINDy-MPC2} demonstrates the tracking position error using a SINDy-MPC and MPC method. The SINDy-MPC has a lower error value than the MPC because our methods can accurately predict a future state using a data-driven model that considers model uncertainty and unknown models, so that the SINDy-MPC achieves better tracking performance. The root mean square error is used to analyze the tracking performance in Table \ref{T21}. The overall RMSE indicates that SINDy-MPC achieves superior tracking performance compared to conventional MPC. Figure \ref{SINDy-Time} demonstrates the thrust value and computation time using SINDy-MPC. When the multirotor meets a static obstacle, an aggressive maneuver is required to avoid collision so that the thrust command is increased quickly. Through a computational control strategy, we can consider a control input constraint such as an upper control input range. The computation time is an important point because this prohibits one from implementing a real system. When the multirotor faces the obstacle, the computation time is increased to find a possible way to evade the obstacle smoothly. The simulation results show that SINDy-MPC can be applied to a real-time application, requiring a lower execution time than a main loop time.

\begin{table}[htb!]
\caption{RMSE of position state using SINDy-MPC and MPC.\label{T21}}
\begin{center}
\begin{tabular}{|c|c|c|c|c|}
\hline
{RMSE [m]} & $x$   & $y$   & $z$ \\ \hline
{ SINDy-MPC } & \textbf{0.8129 }  & \textbf{0.3812}  & \textbf{0.3988}   \\
\hline
{MPC } & 0.9086   & 0.4522  &0.6279     \\ \hline
\end{tabular}
\end{center}
\end{table}

\section{Conclusion} % 이동우
\label{sec6}
In this study, a novel data-driven model predictive control is proposed for a multirotor with payload. The suggested approach integrates sparse identification of nonlinear dynamics and model predictive control to avoid the obstacle during autonomous flight. Using the SINDy, nonlinear dynamics of the multirotor is identified to apply a nominal model for state prediction. Through a simulation, the results show that the proposed method can follow a reference trajectory and avoid collisions under parameter uncertainty and unknown model effects. Our future work will apply the proposed method to the multirotor to confirm the ability of SINDy-MPC to avoid obstacles.

% \bibliographystyle{unsrt}
% \bibliography{llncs}

\end{document}